\pdfoutput=1

\documentclass[11pt]{article}

\usepackage[]{acl}

\usepackage{times}
\usepackage{latexsym}
\usepackage{float}
\usepackage{graphicx}
\usepackage{amsmath}
\usepackage{amssymb}
\usepackage{hyperref}
\usepackage{url}
\usepackage{wrapfig}
\usepackage{caption}
\usepackage{subcaption}
\usepackage{microtype}

\usepackage[T1]{fontenc}

\usepackage[utf8]{inputenc}

\usepackage{microtype}

\title{DecBERT: Enhancing the Language Understanding of BERT\\with Causal Attention Masks}

\author{Ziyang Luo$^{1}$\thanks{\quad equal contribution}, Yadong Xi$^2$\footnotemark[1], Jing Ma$^1$,  Zhiwei Yang$^{1,}$$^3$,\\{\bf Xiaoxi Mao$^2$, Changjie Fan$^2$, Rongsheng Zhang$^2$}\\
  $^1$  {\normalsize Department of Computer Science, Hong Kong Baptist University, Hong Kong SAR, China} \\
  $^2$ {\normalsize Fuxi AI Lab, NetEase Inc., Hangzhou, China} \\
  $^3$ {\normalsize Jilin University, Jilin, China} \\
  \texttt{\normalsize cszyluo@comp.hkbu.edu.hk, majing@hkbu.edu.hk}\\
  \texttt{\normalsize \{xiyadong,zhangrongsheng\}@corp.netease.com}}

\begin{document}
\maketitle
\begin{abstract}

Since 2017, the Transformer-based models play critical roles in various downstream Natural Language Processing tasks. However, a common limitation of the attention mechanism utilized in Transformer Encoder is that it cannot automatically capture the information of word order, so explicit position embeddings are generally required to be fed into the target model. In contrast, Transformer Decoder with the causal attention masks is naturally sensitive to the word order. In this work, we focus on improving the position encoding ability of BERT with the causal attention masks. Furthermore, we propose a new pre-trained language model \textit{DecBERT} and evaluate it on the GLUE benchmark. Experimental results show that (1) the causal attention mask is effective for BERT on the language understanding tasks; (2) our \textit{DecBERT} model without position embeddings achieve comparable performance on the GLUE benchmark; and (3) our modification accelerates the pre-training process and \textit{DecBERT w/ PE} achieves better overall performance than the baseline systems when pre-training with the same amount of computational resources.

\end{abstract}

\section{Introduction}

In recent years, Transformer model proposed by \citet{NIPS2017_3f5ee243} has supplanted the widely-used LSTM \citep{10.1162/neco.1997.9.8.1735} as an indispensable component of many NLP systems. There are two branches of model variant: Transformer Encoder and Transformer Decoder. The Encoder-based Language Models, e.g., BERT \citep{devlin-etal-2019-bert}, RoBERTa \citep{liu2019roberta} and DeBERTa \citep{he2020deberta}, have achieved great success on many natural language understanding benchmarks (e.g. GLUE \citep{wang-etal-2018-glue} and SuperGLUE \citep{wang2020superglue}). The Decoder-based Language Models such as GPT-family \citep{Radford2018ImprovingLU,radford2019language,brown2020language} have shown superior performances on natural language generation. All of them utilize the Multi-Head Self-Attention (\textsc{Mha}) mechanism \citep{NIPS2017_3f5ee243}. Since \textsc{Mha} is designed as an order-invariant mechanism \citep{pmlr-v97-lee19d}, Transformer Encoder without the help of position embeddings should share the same intuitions with the bag-of-word model. On the other hand, in Transformer Decoder, the causal attention masks make the \textsc{Mha} different from that of the Transformer Encoder. Specifically, \citet{tsai-etal-2019-transformer} have proved that \textsc{Mha} with such attention masks is not permutation equivalent, indicating that Transformer Decoder is sensitive to word order.

It is noticed that several studies focus on enriching the position information of BERT to improve the performance of natural language understanding~\cite{dai-etal-2019-transformer,dufter-etal-2020-increasing,he2020deberta,wu2021datransformer,ke2021rethinking}, e.g., introducing extra learnable parameters to trace the word order. Previous analysis also indicate that the lower layers of BERT tend to capture rich surface-level language structural information such as position information~\cite{jawahar-etal-2019-bert}. In this paper, to improve the language understanding of BERT, we propose to enrich the position information in the lower hidden layers instead of introducing extra learnable positional parameters.

To this end, we firstly design analysis experiments to examine the effectiveness of causal attention masks in terms of capturing position information. Then we propose a new pre-trained language model \textit{DecBERT} by adding the causal attention masks into the lower layers of BERT (e.g., the first two layers) to enhance the position encoding ability. In this way, our proposed model is naturally sensitive to word order. Then we pre-train our \textit{DecBERT} as a masked language model, following the same objective as BERT. To verify whether our modification can help BERT trace word order, we also make a comparison with a variant of our \textit{DecBERT} that excludes any position embeddings. The experimental results show that \textit{DecBERT w/o PE} has 77 times (4.59 vs. 353.97) lower valid PPL score than BERT w/o PE and achieves comparable performance with BERT w/ PE on downstream tasks, corroborating the effectiveness of our modification. Furthermore, \textit{DecBERT w/ PE} achieves better performances than BERT on most downstream tasks when pre-training with the same amount of time and computational resources. By analyzing the pre-training process, we find that our modification can also accelerate pre-training.

The contributions of this work are summarised as follows:
\begin{itemize}
    \item We propose a novel pre-train model \textit{DecBERT} utilizing the causal attention masks to enhance language understanding of BERT. 
    \item We show that \textit{DecBERT w/o PE} has comparable performance with BERT w/ PE, indicating that the causal attention masks are effective for modeling word order.
    \item When pre-training with the same amount of time and computational resources, \textit{DecBERT w/ PE} achieves lower validation PPL and better overall performance on GLUE than BERT.
\end{itemize}

\section{Background: Transformer}

Transformer is a neural network model proposed by \citet{NIPS2017_3f5ee243}, which relies on the multi-head self-attention (\textsc{Mha}) mechanism.\\

\noindent\textbf{Input Layer.}\indent Due to the order-invariance of \textsc{Mha} \citep{pmlr-v97-lee19d}, a token embedding is added with a position embedding as the input of Transformer Encoder or Decoder:
\begin{equation}
    h_i=TE(x_i) + PE(i),
\end{equation}
where $x_i$ is a token at the $i^{th}$ position. $TE$ is a token embedding matrix and $PE$ is a position embedding matrix. In the paper of \citet{NIPS2017_3f5ee243}, they use a fixed sinusoidal $PE$:
\begin{equation}
\begin{split}
    PE[i, 2j]=\sin(i/10000^{2j/d_m}),\\
    PE[i, 2j+1]=\cos(i/10000^{2j/d_m}),
\end{split}
\end{equation}
where $j$ is the dimension and $d_m$ is the model size. In the later work, \citet{devlin-etal-2019-bert} choose to use a learnable $PE$ matrix.\\

\noindent\textbf{Multi-head Self-attention (\textsc{Mha}).}\indent \textsc{Mha} takes a sequence of vectors $h=[h_1,h_2,...,h_n]$ as input. Then they are transformed into three different vectors, query (Q), key (K) and value (V), by three linear transformations and passed to the multi-head self-attention (\textsc{Mha}). The computation process of a single head is:
\begin{equation}
    Attention(Q, K, V)=softmax\left(\frac{QK^T}{\sqrt{d_k}}\right)V,
\end{equation}
where $d_k$ is the dimension of a single head. \textsc{Mha} repeats the same process for $h$ heads. The outputs of all heads are concatenated together and passed through a linear projection $W^{O}$ again:
\begin{equation}
\begin{split}
    &H_i=Attention(Q_i,K_i,V_i),\\
    &MHA(Q, K, V)=concat(H_1,...,H_h)W^O.
\end{split}
\end{equation}

\noindent\textbf{Transformer Encoder and Decoder.}\indent An Encoder layer consists of multi-head attention following with a feed-forward network (FFN). The outputs of \textsc{Mha} and FFN are passed through a LayerNorm \citep{ba2016layer} with residual connections \citep{he2015deep}. Then we stack multi-layer to form a Transformer Encoder. The difference between Decoder and Encoder is that Decoder uses the causal attention masks to mask the attention values of the subsequent tokens so that Decoder can only decode tokens relying on the tokens in the past.\footnote{We do not consider the Encoder-Decoder Seq2seq structure with cross attention here. Encoder and Decoder are used independently.}

\begin{figure*}
    \centering
    \includegraphics[height=6.3cm]{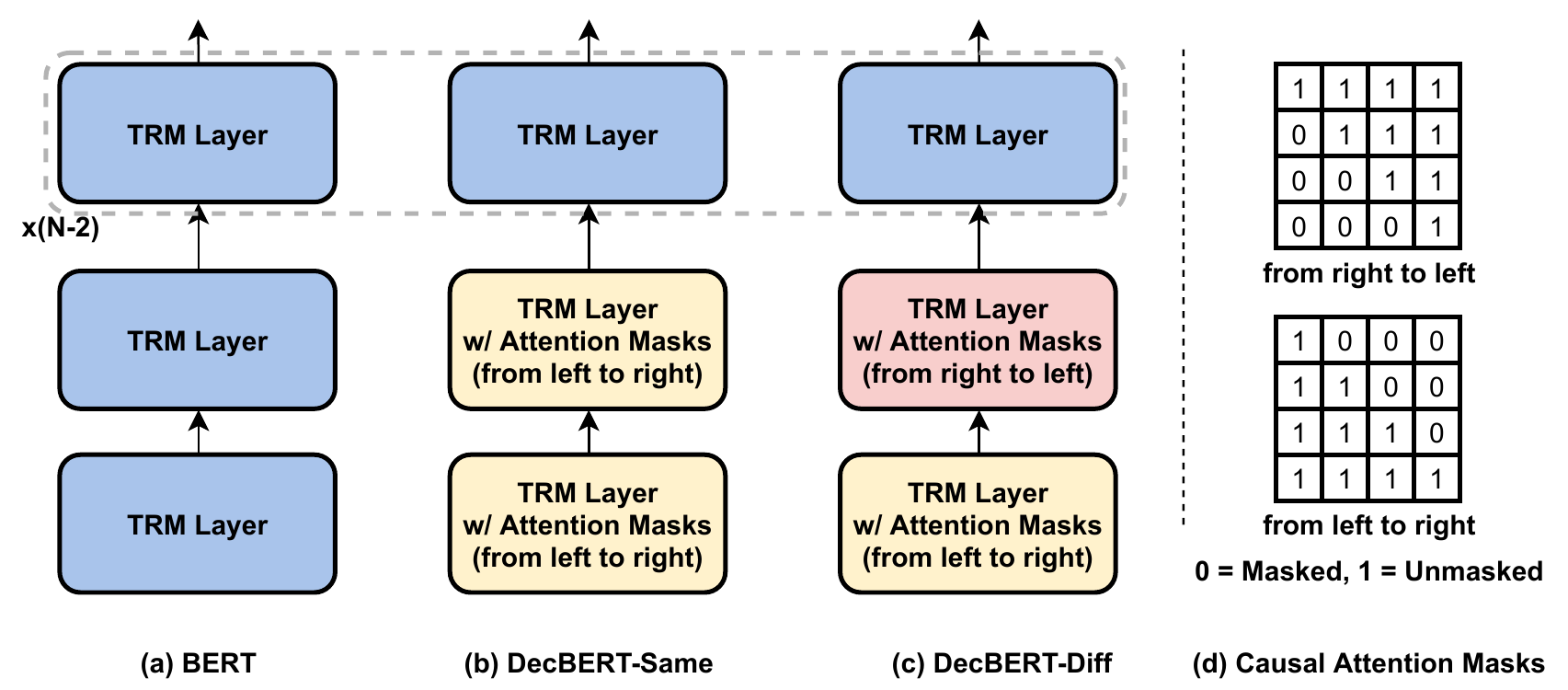}
    \caption{Model structures of BERT and DecBERT. TRM refers to the Transformer layer.}
    \label{fig:DecBERT}
\end{figure*}

\section{Methodology}

In this section, we first analyze the relationship between Transformer Decoder and position embeddings (section \ref{sec:decoder}). Based on this analysis, we inject the causal attention masks into BERT to create our new pre-trained language models, \textit{DecBERT} (section \ref{sec:decbert}).

\subsection{Transformer Decoder and Position Embeddings}\label{sec:decoder}
Previous studies \citep{tsai-etal-2019-transformer} indicate that Transformer Decoder with causal attention masks is sensitive to word order. We wonder whether Transformer Decoder can perform well without position embeddings. We assume that if Transformer Decoder without any position embeddings still retains comparable performance with its counterpart with position embeddings, it will corroborate that the causal attention masks are helpful for Transformer to encode word order.  To this end, we design a straightforward 
experiment of causal language modeling respectively on English and Chinese data as followed.\\

\noindent\textbf{Basic Model.}\indent Our basic model is an 8-layer Transformer Decoder with 768 embedding size, 3072 feedforward layer hidden size, 12 attention heads and GELU activation function \citep{hendrycks2020gaussian}, which is a smaller version of GPT and has 95M trainable parameters for English model and 77.5M for Chinese model.\footnote{The Chinese vocabulary size is smaller than English, so the Chinese model has fewer parameters.} We find that if we use a standard 12-layer GPT, the number of trainable parameters will be higher than the number of tokens in the WikiText-103 dataset. This has a risk to cause over-fitting, so we choose to use an 8-layer model. \\

\noindent\textbf{Data and Training.}\indent We resort to two publicly available wikipedia datasets. The first one is the English WikiText-103 \citep{merity2016pointer}. We train and evaluate our language models on the standard splits of the WikiText-103, which contains 1.8M sentences for training and 3.76k sentences for evaluation. The second one is the Chinese Wikipedia which contains about 9.28M sentences. We randomly select 34k sentences for evaluation and 9.25M for training. We use Fairseq \citep{ott-etal-2019-fairseq} to pre-process all the data into the binary files. All the English data is tokenized by SentencePiece tokenizer \citep{kudo-richardson-2018-sentencepiece}, which is the same as RoBERTa. All Chinese data is tokenized by character.

All models are trained with Fairseq. The training objective is the Causal Language Modeling objective. We use a batch size of 128 and train for 100k steps, optimized by Adam \citep{kingma2017adam}. We also use the polynomial learning rate decay with 10k warmup steps. All models use the same hyper-parameters. We list the details in the Appendix. We use two NVIDIA A100 40GB GPUs to train each model. For the WikiText-103, it costs about 10 hours per model. For the Chinese Wikipedia, it costs about 8.5 hours per model.\\

\begin{table}
    \centering
    \begin{tabular}{lcc}
        \hline
        Transformer Decoder & w/o \textit{PE} & w/ \textit{PE} \\
        \hline
        WikiText-103 & 23.52 & 23.37\\
        Chinese Wiki & 12.96 & 12.75\\
        \hline
    \end{tabular}
    \caption{Transformer Decoders perplexity (PPL) on WikiText-103 and Chinese Wikipedia validation sets. \textit{PE} refers to the learnable position embeddings.}
    \label{tab:PPL}
\end{table}

\noindent\textbf{Results and Discussion.}\indent Table \ref{tab:PPL} presents the perplexity (PPL) scores of Transformer Decoders with or without position embeddings on WikiText-103 and Chinese Wikipedia validation sets. Transformer Decoder w/o PE achieves comparable performance with its counterpart with learnable PE, which is only about 0.2 higher. This result reveals that the additional performance gain brought by position embeddings is small. Only relying on its causal attention masks, Transformer Decoder still can perform well. Combing our experiment and the previous studies \citep{tsai-etal-2019-transformer,Irie_2019}, the causal attention masks can make Transformer sensitive to word order.

\subsection{Our DecBERT Model}\label{sec:decbert}

In section \ref{sec:decoder}, we conclude that Transformer with the causal attention masks is naturally sensitive to word order. Since the position information is inevitable for BERT, we propose to enhance existing BERT model based on causal attention masks.

In this paper, we intend to add the causal attention masks into all or some hidden layers of BERT. In this way, the specific layers with such masks are sensitive to word order by design, which can enhance the position encodings ability of BERT. Such framework can further result in better language understanding performances, e.g., in pre-trained language modeling, casual attention masks were added on all 12 layers of GPT~\cite{Radford2018ImprovingLU}. However, comparing with BERT~\cite{devlin-etal-2019-bert}, we observe that GPT lags behind BERT on almost all downstream tasks.\footnote{Although GPT and BERT are pre-trained with different objectives, the comparison is reasonable due to the same downstream tasks.} 
This is because self-attention mechanism with such masks only consider one-side information flow, it cannot process the input sentence comprehensively and has a high risk of language information loss. Therefore, we can conjecture that it is not effective to use the causal attention masks in all hidden layers. There is a strong need to maintain a balance between the gain of position encoding ability and the loss of language information.

In order to determine which layer(s) should add casual attention masks, we refer to the BERTology work~\cite{jawahar-etal-2019-bert} that conduct comprehensive experiments to analyze and interpret the information captured by each layer of BERT. The experimental results indicate that the lower layers of BERT capture rich language structure information. The position information is also a common structure information, so that we propose to add the causal attention masks into the lower layers (e.g., the first two layers\footnote{We conducted massive experiments by adding the masks in the first, first-two, or first-three layer(s), and the first-two layers achieve the best performance.}) to improve the position encoding ability of BERT. We denote our model as \textbf{DecBERT}. There are two versions of our model, \textbf{DecBERT-Same} and \textbf{DecBERT-Diff}. All of them are 12-layer base size models.

\begin{itemize}
    \item \textbf{DecBERT-Same}: This model has a similar structure as BERT (see Figure~\ref{fig:DecBERT}(a)), but we use the causal attention masks to convert the first two Encoder layers into two Decoder layers with the same direction (from left to right). So the 12-layer model has 10 Encoder layers and 2 Decoder layers, which is shown in Figure~\ref{fig:DecBERT}(b). In this way, the first two layers are naturally sensitive to word order;
    \item \textbf{DecBERT-Diff}: This model is designed to enhance \textbf{DecBERT-Same} to gain more language information from different encoding directions. This model has a same structure as \textbf{DecBERT-Same}, except the second Decoder layer that has the opposite direction (from right to left). Figure~\ref{fig:DecBERT}(c) illustrates the model structure.
\end{itemize}

One would think that \textbf{DecBERT} is similar to Transformer with RNN layer \citep{neishi-yoshinaga-2019-relation}. Note that \textbf{DecBERT} is quite different from it, because \textbf{DecBERT} has similar structure as \textbf{BERT} and both of them require the same amount of computational time, which is much faster than that of Transformer with RNN.

\section{Experiments and Results}

\subsection{Experimental Setup}
Our experiments can be separated into two parts, small-scale pre-training scenario and large-scale pre-training scenario. Since the small-scale pre-training consumes much less time and fewer computational resources, we intend to answer several research questions in this part: 
\begin{itemize}
    \item Can DecBERT \textbf{without} any position embeddings still understand language well?
    \item Can DecBERT \textbf{with} position embeddings outperform BERT?
    \item Is using different directional causal attention masks more helpful than using the same directional?
    \item Why can DecBERT benefit from the causal attention masks, how do such masks affect the pre-training process?
\end{itemize}

For the large-scale pre-training scenario, we intend to examine whether the performance gap between our \textit{DecBERT} and BERT will be diminished after scaling up the pre-training data size and time. Such settings can present a more comprehensive view of whether our modification can benefit the pre-trained language models.

For a fair comparison, we re-implement BERT and pre-train it with the same settings as \textit{DecBERT} in the small-scale and large-scale pre-training. We denote it as \textit{BERT-reImp}.\\

\noindent\textbf{Small-scale Pre-training Scenario.}\indent The pre-training data is the widely-used English Wikipedia Corpus. We randomly select 158.4M sentences for training and 50k sentences for validation. The pre-training objective is the Masked Language Modeling objective. We use a batch size of 256 and pre-train for 200k steps, optimized by Adam. All models use the same hyper-parameters. We list the details in the Appendix. We use four NVIDIA A100 40GB GPUs to pre-train each model, costing about 34.5 hours per model.\\

\noindent\textbf{Large-scale Pre-training Scenario.}\indent Limited by time and computational resources, it is impossible for us to pre-train all models in the small-scale pre-training scenario from scratch in this setting. Thus, we decide to pre-train the best model in the small-scale scenario and the baseline model \textit{BERT-reImp w/ PE} in this part. We use a large amount of pre-training data (around 160GiB\footnote{The details of our pre-training corpus can be seen in the Appendix.}). The batch size is set to 4096 and the pre-training steps are 300k. We pre-train each model with 8 NVIDIA A100 40GB GPUs, costing about 15 days per model. The hyper-parameters details can be also seen in the Appendix.\\

\noindent\textbf{Fine-tuning.}\indent To evaluate the language understanding ability of our models, we fine-tune them with 8 tasks of GLUE benchmark \citep{wang-etal-2018-glue}, including SST-2 \citep{socher-etal-2013-recursive}, QNLI \citep{rajpurkar-etal-2016-squad}, MNLI \citep{williams-etal-2018-broad}, QQP,\footnote{https://www.quora.com/q/quoradata/First-Quora-Dataset-Release-Question-Pairs} MRPC \citep{dolan-brockett-2005-automatically}, CoLA \citep{warstadt2018neural}, RTE\footnote{https://aclweb.org/aclwiki/Recognizing\_Textual\_Entailment} and STS-B \citep{cer-etal-2017-semeval}. All fine-tuning hyper-parameters details are listed in the Appendix. 

\begin{table}
    \centering
    \begin{tabular}{lcc}
        \hline
        Models & w/ \textit{PE} & Valid PPL \\
        \hline
        \multicolumn{3}{l}{\textbf{Baseline}}\\
        BERT-reImp & False & 353.97\\
        BERT-reImp & True & 4.28\\
        \hline
        \multicolumn{3}{l}{\textbf{Ours (w/o position embeddings)}}\\
        DecBERT-Same & False & 4.59\\
        DecBERT-Diff & False & 4.59\\
        \hline
        \multicolumn{3}{l}{\textbf{Ours (w/ position embeddings)}}\\
        DecBERT-Same & True & 4.12\\
        DecBERT-Diff & True & \textbf{4.07}\\
        \hline
    \end{tabular}
    \caption{The validation set perplexity of all models in small-scale pre-training scenario. (w/ \textit{PE} = with learnable position embeddings)}
    \label{tab:roberta-ppl}
\end{table}

\begin{table*}
    \centering
    \begin{tabular}{lccccccccc}
        \hline
        Models & SST-2 & QNLI & QQP & RTE & MNLI-m/mm & MRPC & STS-B & Avg.\\
        \hline
        \multicolumn{7}{l}{\textit{Small-scale pretraining results on the dev sets}}\\
        BERT-reImp & 89.56 & 89.24 & 90.14 & 64.40 & 80.14/80.62 & \textbf{86.60} & 86.22 & 83.37\\
        \hline
        \multicolumn{5}{l}{\textbf{Ours (w/o position embeddings)}}\\
        DecBERT-Same & 89.58 & 89.50 & 90.16 & 62.68 & 79.56/80.42 & 85.88 & \textbf{86.58} & 83.05\\
        DecBERT-Diff & 90.30 & 88.86 & 90.28 & 59.28 & 79.78/80.78 & 86.08 & 86.06 & 82.68\\
        \hline
        \multicolumn{5}{l}{\textbf{Ours (w/ position embeddings)}}\\
        DecBERT-Same & 90.12 & 89.18 & \textbf{90.32} & 64.78 & 80.48/80.64 & 86.24 & 86.34 & 83.51\\
        DecBERT-Diff & \textbf{90.78} & \textbf{89.56} & 90.08 & \textbf{65.98} & \textbf{80.92}/\textbf{81.26} & 85.86 & 86.24 & \textbf{83.84}\\
        \hline
    \end{tabular}
    \caption{Different small-scale pre-training models' performance on the \textbf{dev sets} of GLUE benchmark. All results are averaged over five different random seeds (1, 2, 3, 4 and 5). MNLI-m is the matched version and MNLI-mm is the mismatched version. All tasks except STS-B use accuracy as their evaluation metrics. STS-B uses the Spearman rank correlation. The results are reported as $r\times 100$. \textbf{Bold} indicates the best score for each task.}
    \label{tab:FT}
\end{table*}

\begin{table*}
    \centering
    \begin{tabular}{lcccccccccccc}
        \hline
        Models & SST-2 & QNLI & QQP & RTE & MNLI-m/mm & CoLA & MRPC & STS-B & Avg.\\
        \hline
        \multicolumn{7}{l}{\textit{Large-scale pretraining results on the test sets}}\\
        BERT-reImp & \textbf{94.7} & 91.5 & \textbf{89.4} & 66.5 & 85.9/85.1 & 56.3 & 85.4 & \textbf{86.8} & 82.4\\
        DecBERT-Diff & 94.5 & \textbf{92.0} & 89.3 & \textbf{72.0} & \textbf{86.8/85.5} & \textbf{59.6} & \textbf{86.0} & \textbf{86.8} & \textbf{83.6}\\
        \hline
    \end{tabular}
    \caption{Different large-scale pre-training models' performance on the \textbf{test sets} of GLUE benchmark. All tasks except STS-B and CoLA use accuracy as their evaluation metrics. STS-B uses the Spearman rank correlation. CoLA uses the Matthews correlation coefficient. The results are reported as $r\times 100$. \textbf{Bold} indicates the best score of our models for each task.}
    \label{tab:FT-more-data}
\end{table*}

\subsection{Small-scale Pre-training}

Table~\ref{tab:roberta-ppl} presents the pre-training perplexity scores of all systems on the validation set. Table \ref{tab:FT} shows the performance of different systems on the GLUE benchmark. One can notice that our proposed models achieve lower valid PPL scores and higher overall scores on the downstream tasks.\\

\noindent\textbf{Can DecBERT without any position embeddings still understand language well?}\indent Since the self-attention of Transformer Encoder is order-invariant, the extra position information is inevitable for it to model language. Otherwise, it just becomes a bag-of-word model. From Table \ref{tab:roberta-ppl}, we can find that the valid PPL score of \textit{BERT-reImp w/o PE} is up to 353.97, which is about 82 times higher than its counterpart with position embeddings (4.28), revealing that this bag-of-word model cannot model language well. However, one can notice that \textit{DecBERT} does not have such phenomenon. The valid PPL score of \textit{DecBERT w/o PE} is only about 0.5 higher than \textit{DecBERT w/ PE}. Compared with \textit{BERT-reImp w/o PE}, the causal attention masks can decrease the PPL score by a large margin (from 353.97 to 4.59). After fine-tuning on downstream tasks, Table \ref{tab:roberta-ppl} indicates that \textit{DecBERT-Same/Diff w/o PE} retains the same level performance as \textit{BERT-reImp w/ PE}. These results reveal that \textit{DecBERT} still can understand language well without the help of position embeddings, which is in line with our experimental results in section \ref{sec:decoder}.\\

\noindent\textbf{Can DecBERT with position embeddings outperform BERT?}\indent Table \ref{tab:roberta-ppl} shows that both \textit{DecBERT-Same} and \textit{DecBERT-Diff} have lower validation PPL scores than \textit{BERT-reImp} (w/ PE). After fine-tuning on the downstream tasks, Table \ref{tab:FT} reveals that they also have better performance on most tasks. These results confirm our belief that our models can benefit from the causal attention masks. Such masks enhance the position encoding ability of BERT, leading to better language understanding ability.\\

\noindent\textbf{Is using different directional causal attention masks helpful?}\indent The only difference between \textit{DecBERT-Same} and \textit{DecBERT-Diff} is that we adopt a different directional causal attention mask in the second layer. Table \ref{tab:roberta-ppl} shows that \textit{DecBERT-Diff w/ PE} achieves the lowest validation PPL score (4.07). After fine-tuning on the downstream tasks, it also has the best overall score. These results confirm our belief that \textit{DecBERT} can benefit from using different directional attention masks. Though the first two layers of \textit{DecBERT-Diff} only consider one-side information flow, the model can learn to process different directional information in the first two layers. This design maintains a better balance between the gain of position encoding ability and the loss of language information.\\

\begin{figure}
    \centering
    \includegraphics[height=5cm]{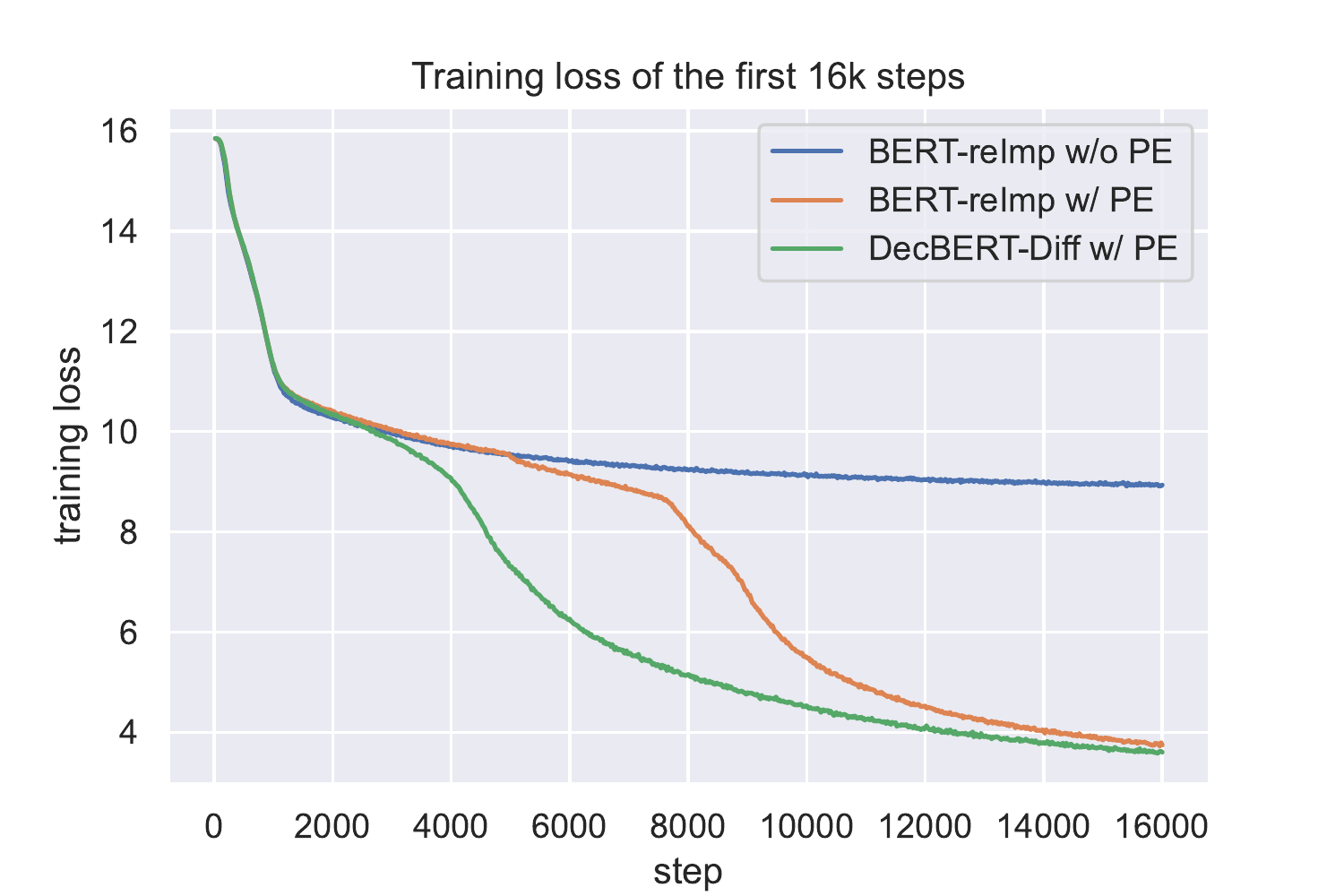}
    \caption{The pre-training loss of the first 16k steps. (Small-scale pre-training)}
    \label{fig:16k}
\end{figure}

\noindent\textbf{Why can DecBERT benefit from the causal attention masks?}\indent The experimental results in the previous part indicate that the causal attention masks can increase the model's position encoding ability. Then such ability leads to better language understanding ability. However, the relation between these two abilities remains unclear. We analyze the pre-training process of our models to give a possible explanation.

\begin{figure}
    \centering
    \includegraphics[height=5cm]{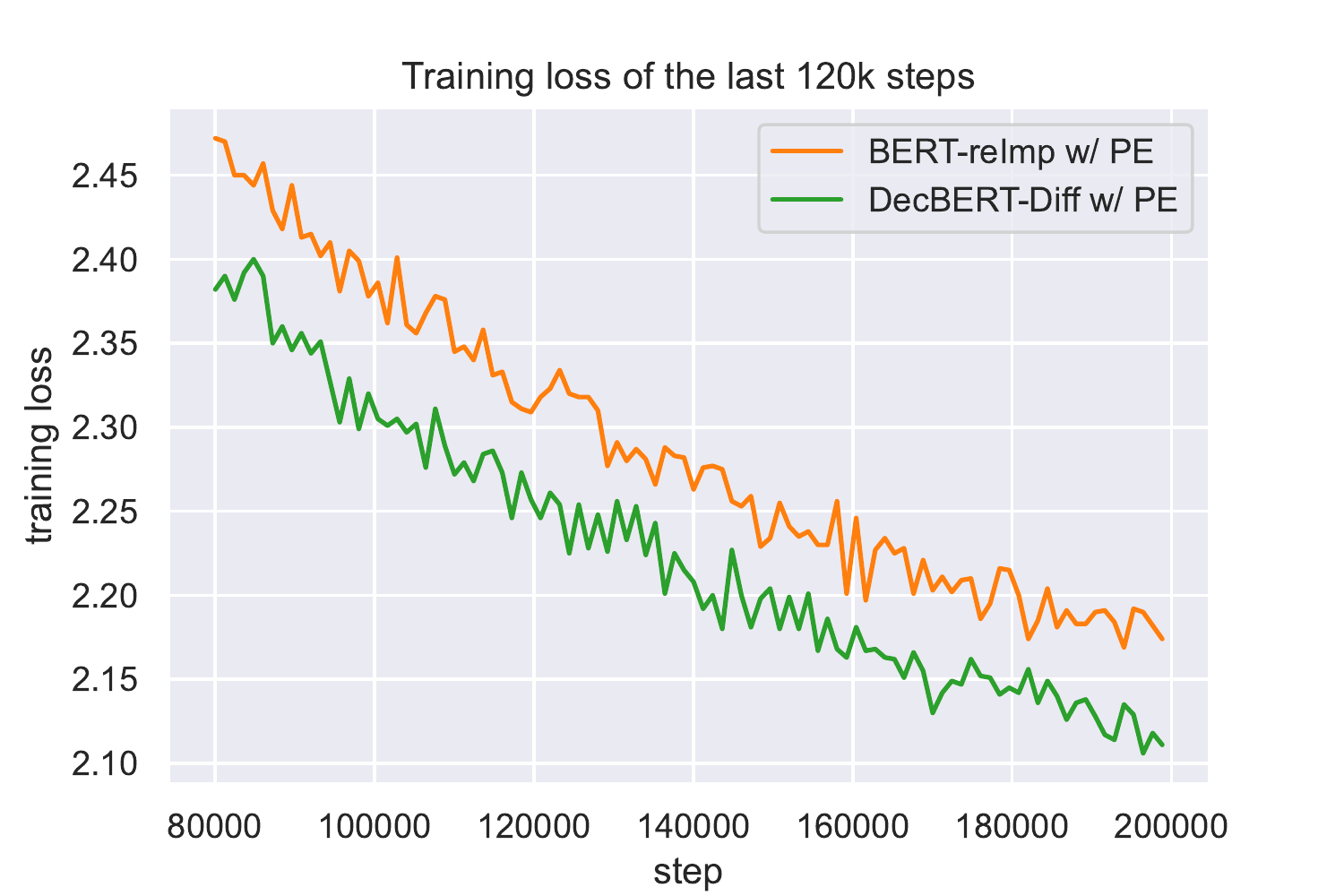}
    \caption{The pre-training loss of the last 120k steps. (Small-scale pre-training)}
    \label{fig:120k}
\end{figure}

Our models' pre-training loss curves are presented in Figure \ref{fig:16k} and \ref{fig:120k}. Since the randomly initialized Multi-head Self-Attention of BERT is a ``balance'' structure without any inductive bias, the model needs to learn suitable position embeddings to trace the word order during pre-training. In Figure \ref{fig:16k}, one can notice that the pre-training process of \textit{BERT-reImp w/ PE} can be divided into four stages: (1) starting stage (0-1000 steps), (2) plateau stage (1000-8000 steps), (3) ``diving'' stage (8000-10000 steps) and (4) convergence stage (10000-final steps). In the starting and plateau stages, \textit{BERT-reImp w/ PE} has almost the same training loss as its counterpart without PE, which indicates that it is still a bag-of-word model and does not know how to make use of the position information. In the ``diving'' stage, the training loss of \textit{BERT-reImp w/ PE} decreases rapidly, while \textit{BERT-reImp w/o PE} starts to converge. This reveals that the word order information becomes more useful for models to understand language in such stage. In the convergence stage, the training loss decreases slowly to the end of the whole pre-training process.

So, how do the causal attention masks affect the pre-training process? The first two layers of \textit{DecBERT} can break the ``balance'' of the multi-head self-attention by design. The position bias from the attention masks makes the first two layers sensitive to word order information. In Figure \ref{fig:16k}, one can notice that the plateau stage of \textit{DecBERT} is shortened (from around 7000 to 3000 steps). This reveals that \textit{DecBERT} does not need to spend as much time as BERT to learn to make use of the position information. It can escape from the bag-of-words sub-optimal point faster. Though the gap between \textit{BERT-reImp w/ PE} and \textit{DecBERT-Diff w/ PE} become smaller in the convergence stage, Figure \ref{fig:120k} indicates that \textit{DecBERT-Diff w/ PE} still has lower training loss in the whole pre-training process.

\subsection{Large-scale Pre-training}

\begin{figure}
    \centering
    \includegraphics[height=5cm]{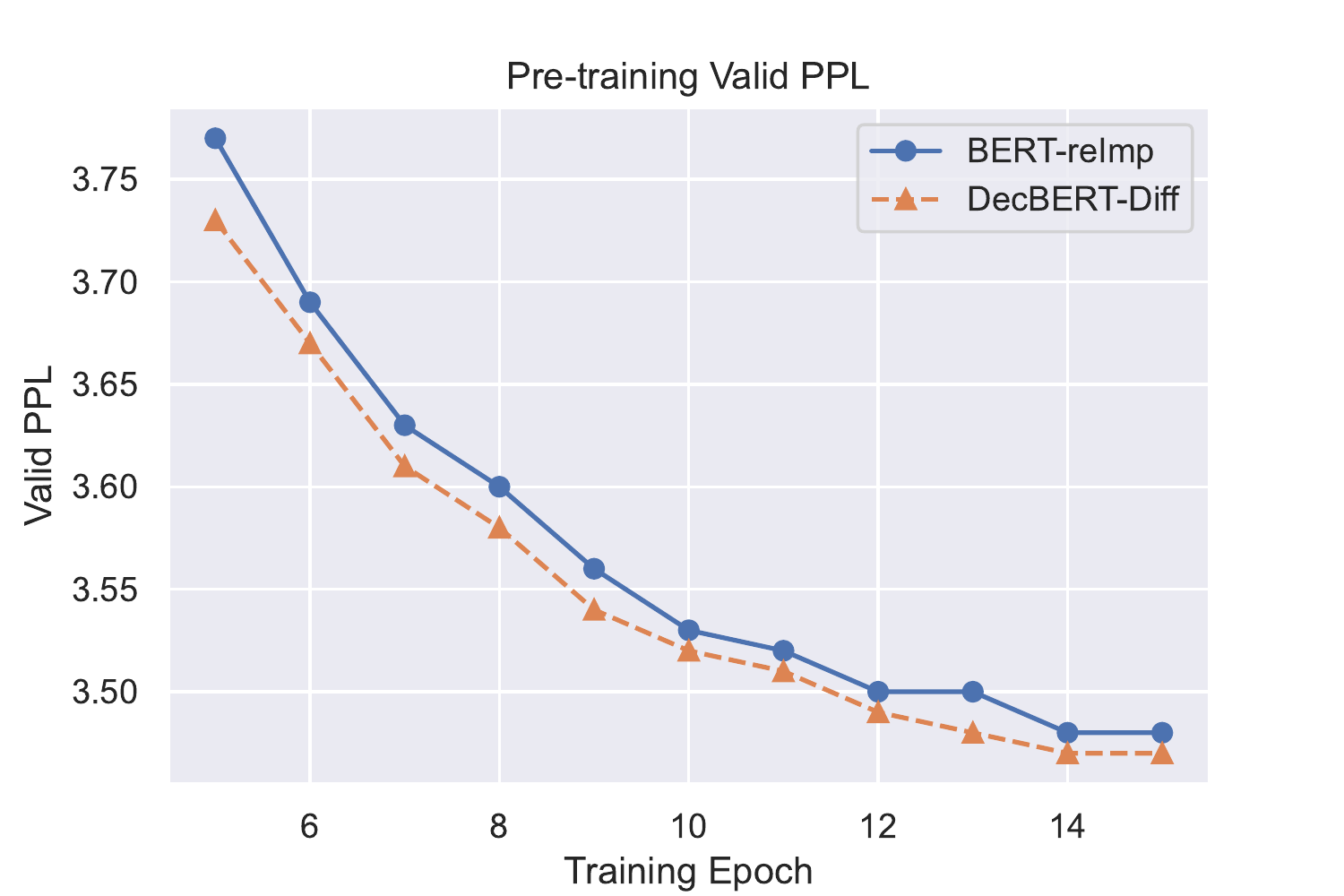}
    \caption{The PPL scores on validation set from epoch 5 to epoch 15 of our models. (Large-scale pre-training)}
    \label{fig:valid_ppl_RoBERTa}
\end{figure}

In the large-scale pre-training scenario, we intend to verify whether our modification still achieves better performance. From Figure \ref{fig:valid_ppl_RoBERTa} and Table \ref{tab:FT-more-data}, one can find that the experimental results are similar to the small-scale pre-training scenario. For the validation PPL, \textit{DecBERT-Diff} achieves lower scores than \textit{BERT-reImp} in the whole pre-training process. Especially, at the $13^{th}$ epoch (265k steps), the valid PPL score of \textit{DecBERT-Diff} is 3.48, which is the same as \textit{BERT-reImp} at the $15^{th}$ epoch (300k steps). This suggests that the pre-training process of \textit{DecBERT-Diff} is about 2 epochs faster than \textit{BERT-reImp}. Combing our previous analysis, one advantage of our modification is that it can accelerate the pre-training process. Comparing the downstream tasks, one can also notice that the performance gap between \textit{DecBERT-Diff} and \textit{BERT-reImp} even becomes larger. The average score is 1.2 points higher.

All results in this part indicate that our modification is effective not only in the small-scale pre-training, but also in the large-scale pre-training. It can accelerate the pre-training process. When pre-training with the same amount of computational resources, our modification can achieve better performance on masked language modeling and downstream tasks.

\subsection{Discussion} 

The analysis and experimental results detailed in the previous sections point out an interesting finding that the pre-training process of BERT can be divided into different stages. A similar phenomenon also can be found in the work of \citet{kovaleva2021bert}. In their work, they find that both scaling factors and biases of the Layer Normalization begin to diverge from their initialization values quickly in the ``diving'' stage. Especially, one/two specific neurons of the biases have larger and larger absolute values. \citet{luo-etal-2021-positional} indicates that such neurons are highly related to the positional information. These complement our possible explanation that in the plateau stage, the model needs to learn suitable position embeddings. Then in the ``diving'' stage, the model learns to adopt such embeddings to better model language. Our \textit{DecBERT} models indicate that breaking the ``balance'' by design can help BERT better capture the position information, which leads to better performance.

\begin{figure}
    \centering
    \includegraphics[height=5cm]{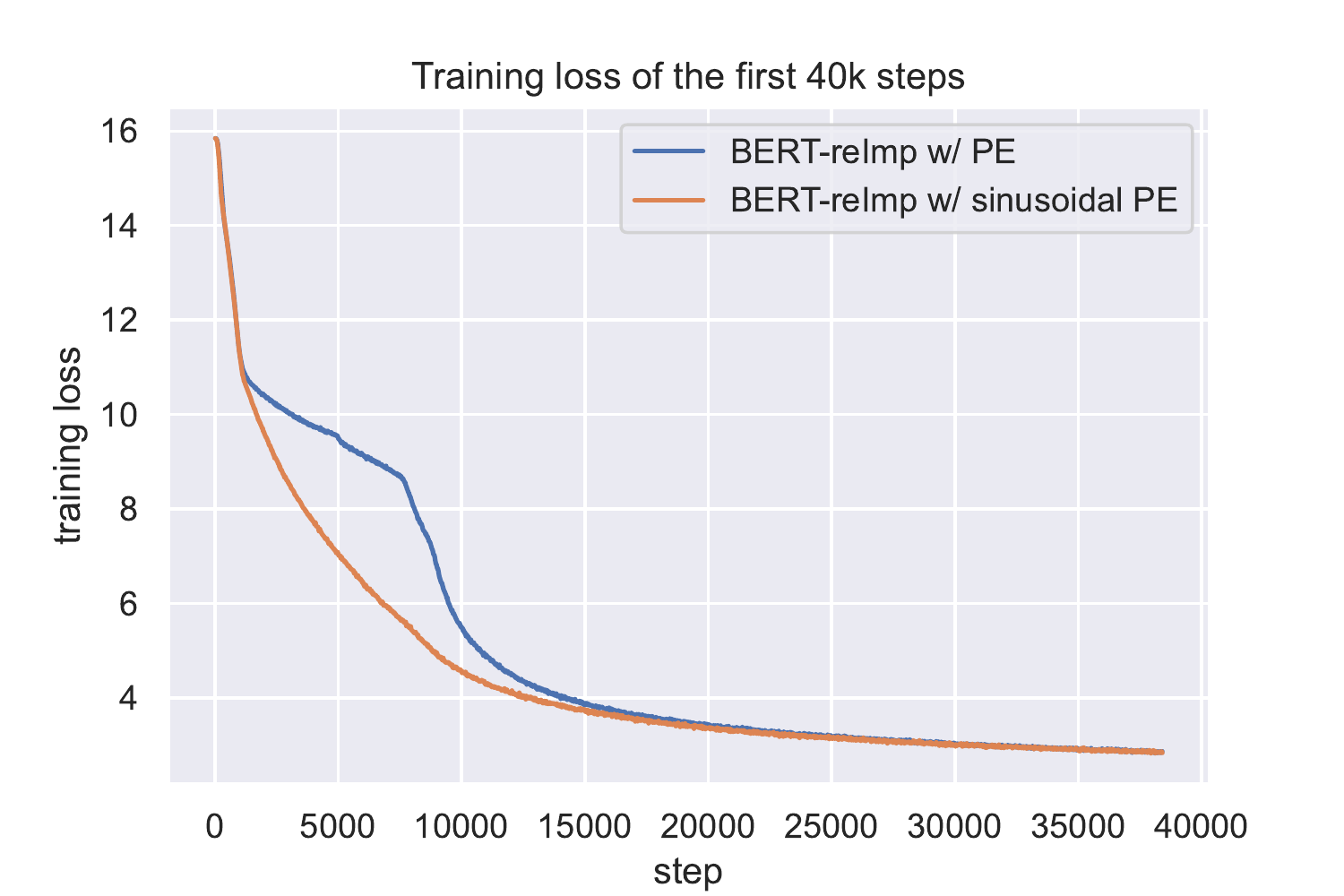}
    \caption{The pre-training loss of the first 40k steps. (Extra small-scale pre-training)}
    \label{fig:sin}
\end{figure}

One would wonder how about the fixed sinusoidal position embeddings. With such embeddings, BERT does not need to learn suitable position embeddings during pre-training. Based on our previous analysis, the plateau stage is possible to disappear. To examine whether such position embeddings are better, we conduct an extra small-scale pre-training experiment. The pre-training loss curve is in Figure \ref{fig:sin}, revealing that the plateau stage indeed disappears. This is in line with our previous results. However, in the convergence stage, we find that BERT with the sinusoidal PE has higher pre-training loss than using the learnable PE. This indicates that the learnable position embeddings are more suitable for BERT.

\section{Related Work}

The previous works \citep{NIPS2017_3f5ee243,shaw-etal-2018-self,huang2018music,dai-etal-2019-transformer,child2019generating} indicate that the self-attention mechanism of Transformer Encoder is permutation equivalent, so it needs to use the position embedding. \citet{tsai-etal-2019-transformer} have proved that Decoder's self-attention is not permutation equivalent, indicating that Decoder is not a bag-of-word model as Encoder, but they do not conduct further analysis on Decoder's position encoding ability. Apart from the analysis, \citet{Irie_2019} train the Transformer Language Models with speech dataset. They find that models without position embeddings have lower perplexity scores. \citet{schlag2021linear} introduce a new Linear Transformer Language Model with fast weight memories \citep{fast1,schlag2021learning}, which has lower perplexity without position encodings on the WikiText-103 dataset.

Furthermore, an explosion of work focuses on proposing a better method to add the position information into the pre-trained language model. \citet{dufter2021position} give a comprehensive introduction of different position encodings methods of Transformer. They divide position encodings into three approaches. One line of such work is to add position embeddings to the input before it is fed to the actual Transformer model \citep{NIPS2017_3f5ee243,shaw-etal-2018-self,devlin-etal-2019-bert,Kitaev2020Reformer,liu2020learning,press2020shortformer,Wang2020Encoding}. The second line of work directly modify the attention matrix \citep{dai-etal-2019-transformer,dufter-etal-2020-increasing,he2020deberta,wu2021datransformer,ke2021rethinking,su2021roformer}. The last one combine the first two approaches together. However, all of them focus on introducing an extra set of parameters to trace the word order. Our work chooses to make use of the causal attention masks.

Most similar to our modification in Section \ref{sec:decbert}, \citet{im2017distancebased} propose a self-attention based model which achieve better performance on SNLI task \citep{bowman-etal-2015-large} without the help of explicit position encodings. However, their models are different from the standard Transformer and use extra local attention masks to control the information flow. With the popularity of the Transformer model in the Computer Vision field, some works propose different methods to make Vision Transformer know word order implicitly \citep{chu2021conditional,yuan2021incorporating,wu2021cvt}, but all of them modify the models with convolution neural network \citep{CNN}.

\section{Conclusion}

In this work, we introduce a new pre-trained model, called \textbf{DecBERT}, adopting the causal attention masks to enhance the language understanding of BERT. We conduct a series of experiments to verify the effectiveness of our models. Experimental results indicate that our proposed models achieve better performance than BERT on most downstream tasks when pre-training with the same amount of data and computational resources. Moreover, our analysis also indicates that our models can accelerate the pre-training process.

\section{Acknowledgments}
We would like to thank Artur Kulmizev and the anonymous reviewers for their excellent feedback. This work is supported by the Key Research and Development Program of Zhejiang Province (No. 2022C01011), HKBU One-off Tier 2 Start-up Grant (Ref. RCOFSGT2/20-21/SCI/004) and HKBU direct grant (Ref. AIS 21-22/02).

\bibliography{anthology,custom}
\bibliographystyle{acl_natbib}

\appendix
\section{Hyper-parameters Details}\label{app:hyper}

\begin{table}[H]
    \centering
    \begin{tabular}{lc}
        \hline
        \textbf{Hyper-parameter} & w/ or w/o PE \\
        \hline
        Number of Layers & 8\\
        Hidden size & 768\\
        FNN inner hidden size & 3072\\
        Attention Heads & 12 \\
        Attention Head size & 64 \\
        Dropout & 0.1\\
        Warmup Steps & 10k\\
        Max Steps & 100k\\
        Learning Rates & 5e-5\\
        Batch Size & 128\\
        Weight Decay & 0.001\\
        Learning Rate Decay & Polynomial\\
        Adam $\epsilon$ & 1e-6\\
        Adam $\beta_1$ & 0.9\\
        Adam $\beta_2$ & 0.998\\
        Gradient Clipping & 0.1\\
        Random Seed & 1\\
        \hline
    \end{tabular}
    \caption{Hyper-parameters for pre-training the Transformer Decoder Causal Language Models.}
    \label{tab:decoder}
\end{table}

\begin{table}[H]
    \centering
    \begin{tabular}{lc}
        \hline
        \textbf{Hyper-parameter} & BERT/DecBERT \\
        \hline
        Number of Layers & 12\\
        Hidden size & 768\\
        FNN inner hidden size & 3072\\
        Attention Heads & 12 \\
        Attention Head size & 64 \\
        Dropout & 0.1\\
        Warmup Steps & 10k\\
        Max Steps & 200k\\
        Learning Rates & 1e-4\\
        Batch Size & 256\\
        Weight Decay & 0.01\\
        Learning Rate Decay & Polynomial\\
        Adam $\epsilon$ & 1e-6\\
        Adam $\beta_1$ & 0.9\\
        Adam $\beta_2$ & 0.98\\
        Gradient Clipping & 0.5\\
        Random Seed & 1\\
        \hline
    \end{tabular}
    \caption{Hyper-parameters for pre-training the BERT and DecBERT (\textbf{small-scale pre-training}).}
    \label{tab:encoder}
\end{table}

\begin{table}
    \centering
    \begin{tabular}{lc}
        \hline
        \textbf{Hyper-parameter} & BERT/DecBERT \\
        \hline
        Number of Layers & 12\\
        Hidden size & 768\\
        FNN inner hidden size & 3072\\
        Attention Heads & 12 \\
        Attention Head size & 64 \\
        Dropout & 0.1\\
        Warmup Steps & 24k\\
        Max Steps & 500k\\
        Learning Rates & 3e-4\\
        Batch Size & 4096\\
        Weight Decay & 0.01\\
        Learning Rate Decay & Tri\_stage\\
        Adam $\epsilon$ & 1e-6\\
        Adam $\beta_1$ & 0.9\\
        Adam $\beta_2$ & 0.98\\
        Gradient Clipping & 2.0\\
        \hline
    \end{tabular}
    \caption{Hyper-parameters for pre-training the BERT and DecBERT (\textbf{large-scale pre-training}).}
    \label{tab:encoder-large-scale}
\end{table}

\begin{table*}
    \centering
    \begin{tabular}{lcccccccc}
        \hline
        \textbf{Hyper-parameter} & MNLI & QNLI & QQP & RTE & SST-2 & MRPC & STS-B & CoLA\\
        \hline
        Learning Rates & 1e-5 & 1e-5 & 1e-5 & 2e-5 & 1e-5 & \{1e-5, 2e-5\} & 2e-5 & 1e-5\\
        Weight Decay & 0.01 & 0.01 & 0.01 & 0.01 & 0.01 & 0.01 & 0.01 & 0.01\\
        Batch Size & 32 & 32 & 32 & 16 & 32 & 16 & 16 & 16\\
        Warmup Steps & 7432	& 1986 & 28318 & 122 & 1256 & 137 & 214 & 320\\
        Max Steps & 123873 & 33112 & 113272 & 2036 & 20935 & 2296 & 3598 & 5336\\
        Adam $\epsilon$ & 1e-6 & 1e-6 & 1e-6 & 1e-6 & 1e-6 & 1e-6 & 1e-6 & 1e-6\\
        Adam $\beta_1$ & 0.9 & 0.9 & 0.9 & 0.9 & 0.9 & 0.9 & 0.9 & 0.9\\
        Adam $\beta_2$ & 0.999 & 0.999 & 0.999 & 0.999 & 0.999 & 0.999 & 0.999 & 0.999\\
        Gradient Clipping & 1.0 & 1.0 & 1.0 & 1.0 & 1.0 & 1.0 & 1.0 & 1.0\\
        \hline
    \end{tabular}
    \caption{Hyper-parameters for fine-tuning all models on downstream tasks. All models use the polynomial learning rate decay. Most of the hyper-parameters are recommended by Fairseq \url{https://github.com/pytorch/fairseq/tree/main/examples/roberta/config/finetuning}.}
    \label{tab:FT-hyper}
\end{table*}

\section{The details of the large-scale pre-training corpus}\label{app:large-scale-pre-train}

The first part is the same as BERT. We use the English wikipedia dump (about 17 GiB) and the bookcorpus \citep{zhu2015aligning} (about 4 GiB). The second part is based on the Pile dataset \citep{gao2020pile}, which is a large datasets with 800 GiB diverse text data. We randomly extract 64 GiB data from the Pile-cc block, 35 GiB data from the OpenWebText2 block and 43 GiB data from the Books3 block. The overall size of all data is about 163 GiB.

\end{document}